\documentclass[journal]{IEEEtran}
\usepackage{times}  
\usepackage{helvet} 
\usepackage{courier}  
\usepackage[hyphens]{url}  
\usepackage{graphicx} 
\usepackage{graphicx}  

\usepackage{lineno}
\usepackage{subfigure}
\usepackage{float}
 \usepackage{graphicx}

\usepackage{CJK}
\usepackage[utf8]{inputenc}
\usepackage[utf8]{luainputenc}
\usepackage{multirow}
\usepackage{array}
\usepackage{color}
\usepackage{bm}
\usepackage{cite}
\usepackage{multirow}
\usepackage{bm}
\usepackage{amsmath}
\usepackage{amsfonts}
\ifCLASSINFOpdf
\else
\fi

\begin{document}

\title{MTNet: A Multi-Task Neural Network for On-Field Calibration of Low-Cost Air Monitoring Sensors}

\author{Anonymous}
\author{Haomin Yu, Yangli-ao Geng, Yingjun Zhang, ~\IEEEmembership{Member,~IEEE}, \\Qingyong~Li,~\IEEEmembership{Member,~IEEE,} Jiayu Zhou,~\IEEEmembership{Member,~IEEE}  \\\thanks{Haomin Yu, Yangli-ao Geng, Yingjun Zhang, Qingyong~Li are with the Beijing Key Lab of Traffic Data Analysis and Mining, Beijing Jiaotong University, Beijing 100044, China.}
\thanks{Jiayu Zhou is with Department of Computer Science and Engineering, Michigan State University, Michigan 48824, America.}
\thanks{Corresponding author: Qingyong Li; e-mail: liqy@bjtu.edu.cn.}}

\maketitle

\begin{abstract}
 The advances of sensor technology enable people to monitor air quality through widely distributed low-cost sensors.
However, measurements from these sensors usually encounter high biases and require a calibration step to reach an acceptable performance in down-streaming analytical tasks. Most existing calibration methods calibrate one type of sensor at a time, which
we call single-task calibration. Despite the popularity of this single-task schema, it may neglect interactions among calibration tasks of different sensors, which encompass underlying information to promote calibration performance.
In this paper, we propose a multi-task calibration network (MTNet) to calibrate multiple sensors (e.g., carbon monoxide and nitrogen oxide sensors) simultaneously, modeling the interactions among tasks. MTNet consists of a single shared module, and several task-specific modules. Specifically, in the shared module, we extend the multi-gate mixture-of-experts structure to harmonize the task conflicts and correlations among different tasks; in each task-specific module, we introduce a feature selection strategy to customize the input for the specific task. These improvements allow MTNet to learn interaction information shared across different tasks, and task-specific information for each calibration task as well. We evaluate MTNet on three real-world datasets and compare it with several established baselines. The experimental results demonstrate that MTNet achieves the state-of-the-art performance.


\end{abstract}



\definecolor{limegreen}{rgb}{0.2, 0.8, 0.2}
\definecolor{forestgreen}{rgb}{0.13, 0.55, 0.13}
\definecolor{greenhtml}{rgb}{0.0, 0.5, 0.0}

%
\IEEEpeerreviewmaketitle

\section{Introduction}
%
%
%
%
\IEEEPARstart{T}{he} atmospheric environment affects many aspects of human lives \cite{castell2017can}, and it is thus particularly significant to monitor air quality. To monitor air quality, governments have set up many static stations \cite{yu2020airnet,wang2020category,yi2018deep}, which can provide reliable air pollutant measurement results. However, high installation and maintenance costs restrict their coverage area. With the continuous advancement of sensor technologies, low-cost air monitoring sensors \cite{yu2011micro,he2017novel,yu2020airnet} are widely deployed to finely monitor air quality in various regions, whereas their measurements are not accurate enough \cite{ferrer2019comparative,park2019energy}. In particular, when a sensor is deployed in the field environment, its sensitivity is affected by uncontrollable environmental factors and decay over time \cite{maag2018survey,clements2017low,barcelo2019self}.


To achieve dense monitoring while ensuring the accuracy of pollutant concentration readings, it is necessary to calibrate the readings of low-cost sensors \cite{de2009co}. 
 In order to solve the calibration problem, many scholars have developed and studied low-cost sensor calibration methods.
According to whether there are high-fidelity reference measurements, the calibration method can be divided into three categories: blind calibration \cite{wang2009blindly, stankovic2015distributed, balzano2007blind}, semi-blind calibration \cite{tan2013system,  kumar2013automatic}, and non-blind calibration (reference-based calibration) \cite{yu2020airnet , wang2020category}.
 \begin{figure}
  \centering
  \includegraphics[width=3.3in]{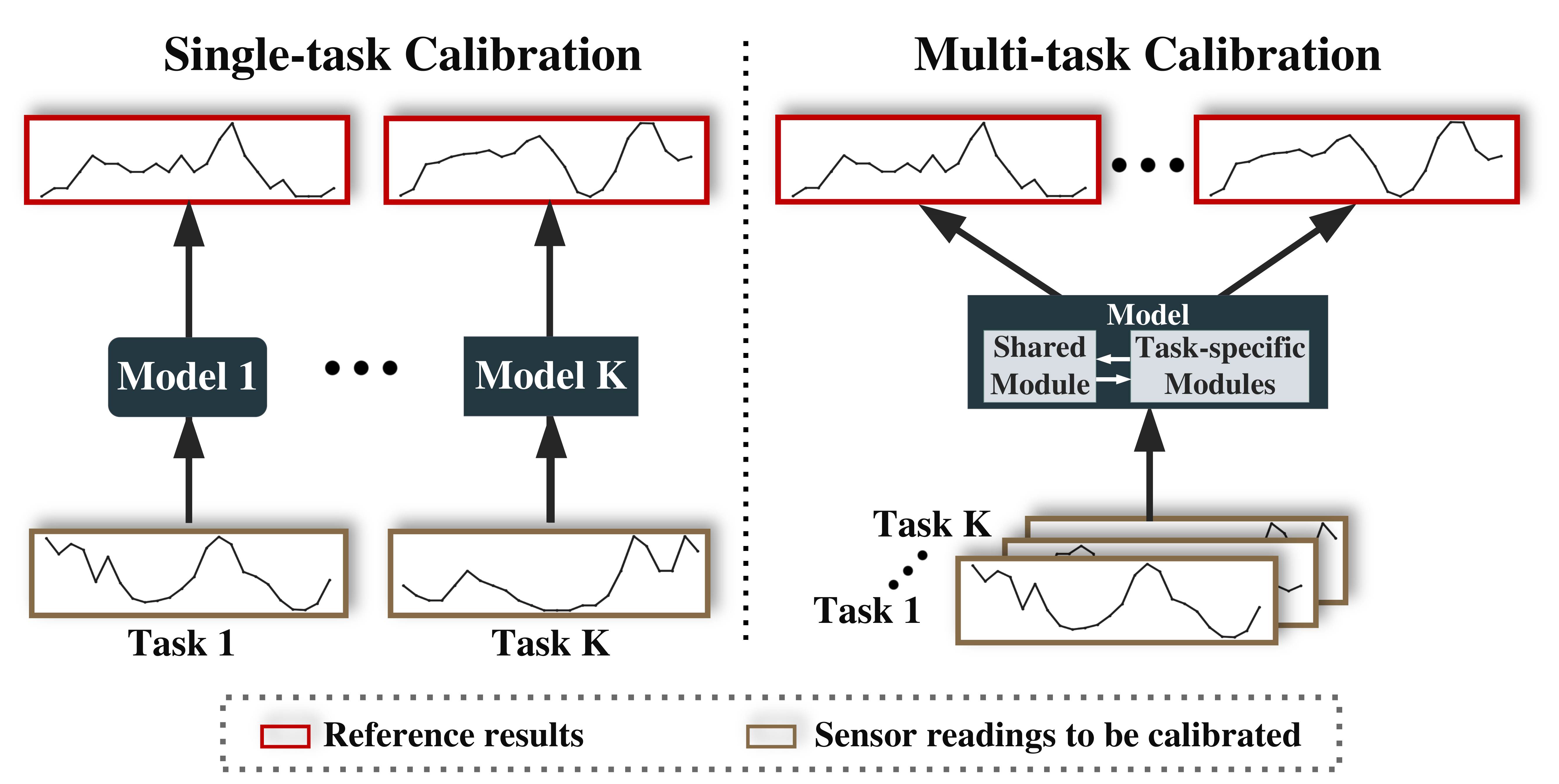}\\
  \caption{A comparison of sketch maps between single-task calibration and multi-task calibration.}
  \label{fig:ske}
\end{figure}
In our task, we focus on reference-based field calibration, meaning that the low-cost sensors are calibrated to reliable reference instrument measurements. The existing reference-based methods mainly follow a single-task calibration schema, utilizing machine learning and deep learning techniques to overcome cross interference\footnote{Cross interference means that the sensor's readings depend not only on the gas being monitored, but also on other air pollutants and environmental factors.} \cite{korotcenkov2013handbook}, sensor drift\footnote{Drift phenomenon usually happens when sensors are
affected by aging and impurities, and it could indirectly cause big errors of low-cost sensors.} \cite{barcelo2019self, kizel2018node}, and limited available features.

Although the single-task schema has achieved some successes in field calibration, it still faces several
limitations. On the one hand, the single-task calibration schema is inefficient in terms of deployment and maintenance since it independently establishes different models for different low-cost sensors. For example, as illustrated in Fig. \ref{fig:ske}, CO, NO$_2$ and NO$_x$ are the three most common gases  of air quality monitoring. To calibrate the sensors of these gases, single-task learning needs to deploy and maintain three models, while multi-task learning only needs to deploy and maintain one model. On the other hand, single-task learning fails to fully exploit the feature interaction among sensors.  Existing researches \cite{seinfeld2016atmospheric,chan2017biomass,duan2019observation,lee2019vehicle} show that carbon monoxide and nitrogen oxides share similar emission sources such as biomass combustion, fossil fuel combustion, and vehicle exhaust emission. Consequently, the concentration curves of carbon monoxide and nitrogen oxides often display similar trends (more details will be presented in Section II.B). Thus, it is worth to exploit this correlation to jointly promote the calibration performance for these pollutants.

To alleviate the aforementioned limitation of single-task calibration, we formulate calibration tasks of multiple sensors into a multi-task calibration framework, calibrating multiple sensors simultaneously. This multi-task formulation enables us to establish more powerful model, whereas it raises several new challenges (will be detailed in Section III). To address these challenges, we propose  multi-task neural network (MTNet) for jointly calibrating sensors of carbon monoxide and nitrogen oxides. The proposed framework consists a shared module and multiple task-specific modules. In the shared module, we extend the multi-gate mixture-of-experts \cite{ma2018modeling} structure to coordinate the task conflicts and correlations among different tasks. In each task-specific module, we introduce a feature selection strategy to customize the input for the specific task. We show that MTNet achieves higher calibration performance on almost all tasks over single-task baselines. In brief, this work has the following contributions. 
\begin{itemize}
  \item We formulate multiple low-cost sensor calibration tasks as a multi-task learning problem instead of the traditional single-task calibration framework. The new formulation considers the interactions among different tasks and supports to calibrate several sensors simultaneously. 
  \item We extend the multi-gate mixture-of-experts architecture and equip MTNet with it to coordinate the correlations and conflicts among different calibration tasks.
  \item  Experimental results on three real datasets show that our model is suitable for joint calibration of carbon monoxide and nitrogen oxide sensors, and achieves an overall performance improvement compared to existing baselines. 

\end{itemize}
The remainder of this paper is organized as follows. Section II presents the data description and problem reformulation. Section III gives the details of the proposed model. Section IV reports the experimental settings and evaluates the performance of the proposed method. Section V provides the conclusions.
 \begin{figure}
  \centering
  \includegraphics[width=3.0in]{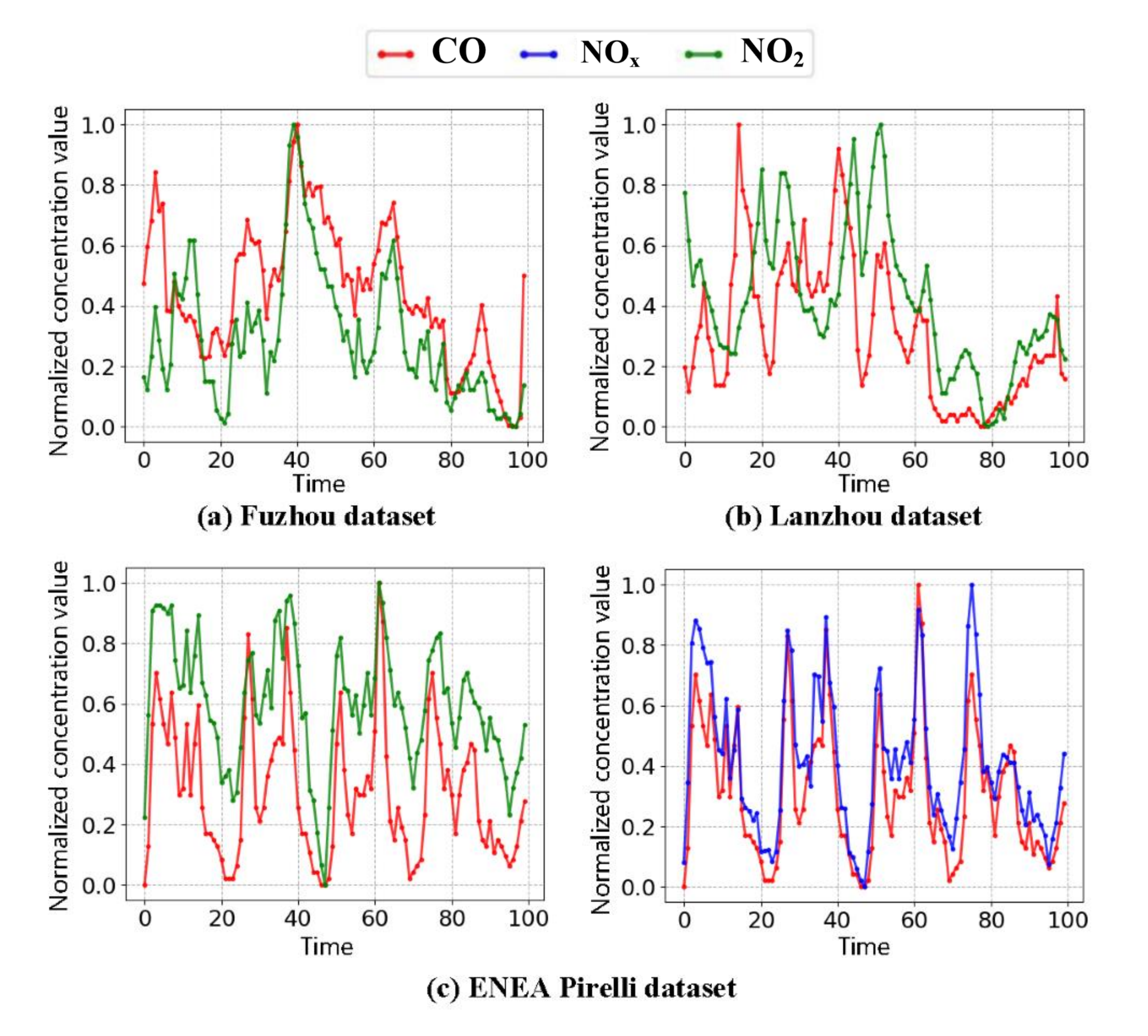}\\
  \caption{The four sub-figures demonstrate the reference concentration curves of carbon monoxide and nitrogen oxide sensors in (a) Fuzhou dataset, (b) Lanzhou dataset and (c) ENEA Pirelli dataset. We randomly select a continuous period from the total time interval for each dataset and draw the corresponding curves over this period.  All the concentrations have been linearly normalized into $[0, 1]$ to eliminate the difference in data scale of different pollutants.}
  \label{fig:tongyuan}
\end{figure}

\begin{figure*}
  \centering
  \includegraphics[width=6.2in]{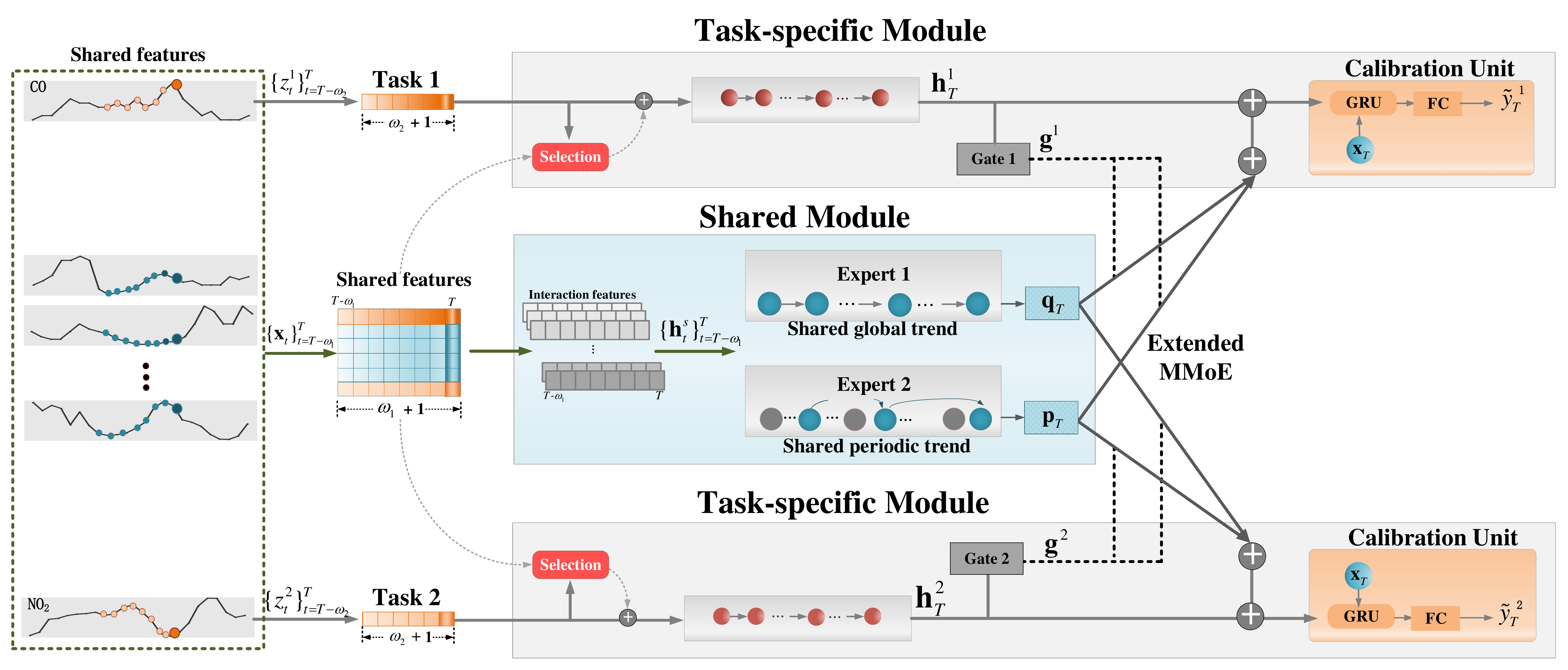}\\
  \caption{The architecture of MTNet. The shared module acquires global and period trend information from shared features. The task-specific module extract task-related information from the corresponding task-specific feature and integrates it with the acquired trend information, generating calibration results. 
  To clearly show our proposed architecture, we utilize joint calibration of two tasks (such as CO and NO$_2$ calibration tasks) as an illustration.}
  \label{fig:random}
\end{figure*}

\section{Preliminary}
\subsection{Data Description}
In our scenario, we aim to calibrate low-cost sensor readings to reference pollutant concentration measurements. In real-world field calibrations, there are other variables (features) available for assisting calibration. Next, we will describe information and notations of sensor-related observation features and reference measurements. Note that the detailed feature information will be presented in the experiment section.

\subsubsection{Observation Features}
These features comprise two parts: task-specific features and shared features.

\begin{itemize}
  \item The task-specific features consist of sensor readings we aim to calibrate. For example, when calibrating a CO sensor, its task-specific feature is the readings of the CO sensor. We use $z_t^k \in \mathbb{R}$ to symbolize the task-specific feature (sensor reading to be calibrated) at time $t$ for the $k$-$th$ sensor.
  \item The shared features are shared by all calibration tasks and provide supplementary information beyond task-related features. These features comprise readings of all sensors (including those to be calibrated), climate features and empirical features \cite{yu2020airnet,yu2020deep}. We utilize symbolic representation ${{\bf{x}}_t} = \{ x_t^s\} _{s = 1}^{{S}} \in {\mathbb{R}^{{S}}}$ to represent shared features at time $t$, where $S$ represents the number of shared features.

\end{itemize}

\subsubsection{Reference Measurements}
The reference measurements acquired from static stations serve as labels in our calibration task. Note that each reference measurement uniquely corresponds a task-specific feature. Thus, for the reference result corresponding to the $k$-$th$ sensor at time $t$, the notation description is denoted by $y_t^k$.

\subsection{Problem Reformulation}
The calibration task can be viewed as a regression problem, i.e., learning a regression function $F_\Theta$ from observation features to reference measurements, where $F_\Theta$ is characterized by the parameter set $\Theta$. Typically, early works \cite{spinelle2015field,spinelle2017field} formulate the calibration task as a multivariate linear regression problem.
Specifically, to calibrate the $k$-th senor reading at time $T$,  they learn a linear regression function from the feature vector $\mathbf{x}_T$ to a scalar $\tilde{y}_T^k$:

\begin{equation}
\label{a11}
 \begin{aligned}
\tilde y_T^k = {F_\Theta }({{\bf{x}}_T}),
\end{aligned}
\end{equation}
where $\tilde{y}_T^k$ is expected to approximate $y_T^k$. This formulation is intuitive and easy to implement. However, it handles the features of different time independently and fails to mine the temporal relationship within the features.

In recent years, researchers managed to introduce temporal dependencies into the formulation.
They introduce a time window of length $\omega_1$ and fetch the feature series within this window to constitute the input. Thus the calibration task can be regarded as a time series regression problem:
\begin{equation}
\label{a22}
 \begin{aligned}
\tilde y_{T}^k &=  {F_\Theta }(\{ {{\bf{x}}_t}\} _{t = T - {\omega _1}}^T).
 \end{aligned}
\end{equation}
The formulation (\ref{a22}) takes  advantage of temporal links within features and thus approaches a higher calibration performance than (\ref{a11}). However, it processes every features  indiscriminately, ignoring variable contributions of different features to the target sensor reading.

Considering that carbon monoxide and nitrogen oxides share similar emission sources, potential links may exist between them. To validate this idea, we draws the concentration curves of these two types of pollutants, as shown in Fig. \ref{fig:tongyuan}. We observe that the two curves in each sub-figure exhibit high consistency, especially at the peak position, suggesting a strong correlation between carbon monoxide concentration and nitrogen oxide concentration. The above observations motive us to introduce the novel task-specific features $z_t^k$ and reformulate the calibration task as a multi-task regression problem:
\begin{equation}
\label{a33}
 \begin{aligned}
\left( \begin{array}{l}
\tilde y_T^1\\
 \vdots \\
\tilde y_T^K
\end{array} \right) 
 &= {F_\Theta }\left( {\{ {{\bf{x}}_t}\} _{t = T - {\omega _1}}^T, \{z_t^1,\dots,z_t^K\}_{t=T-\omega_2}^T} \right),
\end{aligned}
\end{equation}
where $K$ denotes the number of calibration tasks and $\omega_2 + 1$ is the length of the time window for task-specific features. Compared with (\ref{a22}), the proposed formulation (\ref{a33}) exhibits two advantages. First, it distinguishes task-specific features from shared features and enhances contributions of task-related information. Second, it jointly handles multiple calibration tasks and mines the underlying interaction among the different tasks.

\section{Model}
To better illustrate the underlying motivations for our model design, we first summary the challenges about multi-task framework:

\begin{itemize}
\item How to customize the feature set for each task.  As the multi-task model simultaneously handles several calibration tasks, different tasks (sensor readings) are affected by different factors, which impose different requirements on input features. Thus, the model is expected to dynamically extract relevant features for each specific task.
    \item How to balance the correlations and conflicts of multiple tasks in a model. Different tasks pursue different targets, leading the model to different directions. Therefore, it requires the multi-task model to coordinate the task correlations and conflicts. The correlations among tasks are due to the consistency of the external environment and cross-interference phenomenon. Being able to extract and leverage such relationship from learning tasks is expected to significant improve the model performance. However, each task has its own specific characteristics. Thus, it is necessary to overcome task conflicts when multiple tasks are trained and optimized together. Otherwise it may cause performance degradation due to neglect transfer.
                    \end{itemize}

Taking the above challenges into consideration, we propose multi-task neural network (MTNet). As illustrate in Fig. \ref{fig:random}, MTNet consists of two types of modules: the shared module and the task-specific module. The former acquires global and period trend information from shared features; the latter extracts task-related information from the corresponding task-specific feature and integrates it with the acquired trend information. The final calibration results are generated based on the fusion of information extracted by the two types of modules. We detail these two types of modules in the next two subsections. 


\subsection{Shared Module}
This module first extracts interaction features from the input shared feature set. Then, based on the extracted features, global and periodic trend information are acquired by two gated recurrent unit components. We detail the two stages below.
\subsubsection{Interaction Feature}
Considering the cross interference of low-cost sensors, we extract interaction features among multiple shared features. Cross interference means that the readings of sensor is not only affected by the monitored gas, but also interfered by other environmental factors \cite{barcelo2019self,korotcenkov2013handbook}. To overcome this challenge, Yu et al. \cite{yu2020airnet,yu2020deep} proposed to use a single-layer convolution operation to capture interaction characteristics and alleviate cross interference. This operation comprises multiple convolution kernels, with the length and width of the kernels equaling to the feature dimension and local sliding windows, respectively. In our work, as multiple sensors are calibrated simultaneously, a two-layer convolution is utilized  to mine deeper interaction features. Specifically, we denote the two-layer convolution by $Conv( \cdot )$ and formulate this process as

\begin{equation}
\label{eq1}
\{{\bf{h}}_t^s\} _{t = T - {\omega _1}}^{T} = Conv(\{ {{\bf{x}}_t}\} _{t = T - {\omega _1}}^T),
\end{equation}
where $\{{\bf{h}}_t^s\} _{t = T - {\omega _1}}^T$ represents the yielded interaction features.

\subsubsection{Sequence Trend Feature}
Based on the interaction features, MTNet extracts the temporal features within the time series input. Specifically, we intend to capture two types of temporal features: the global trend and the periodic trend.


Given that the gated recurrent unit (GRU) \cite{cho2014learning} can adaptively capture the time dependence of features at different times and overcome vanishing gradient problems \cite{bengio1994learning}, we utilize GRU to extract trend features. Specifically, we feed the interaction features $\{{\bf{h}}_t^s\} _{t = T-\omega_1}^{T}$ into a GRU and return the hidden state of the last time ${\bf{q}}_T$ as the global trend feature:

\begin{equation}
\label{eq2}
{{\bf{q}}_T} = GR{U^{{\rm{global}}}}(\{{\bf{h}}_t^s\} _{t = T - {\omega _1}}^T).
\end{equation}

 The periodic trend is motivated by the fact that concentration of air pollutants pose strong relevance to periodic social activities. For example, the air pollution in the morning and evening peaks is usually more serious than that in other periods. We utilize a GRU-based recurrent-skip component \cite{lai2018modeling,yu2020deep} to extract periodic trend feature. This process is formulated as:
\begin{equation}
\label{eq3}
\begin{array}{l}
{{{\bf{p}}}_T} = GR{U^{\rm{periodic}}}(\{{\bf{h}}_t^s\}_{T-p\cdot {l}}\}_{l=0}^{L}),
\end{array}
\end{equation}
where ${{\bf{p}}_T}$ represents the hidden states at time $T$, which we call periodic trend feature.  $p$ denotes the length of a period and $L$ is the number of periods (i.e., $L$ is equal to $\lfloor \frac{{{\omega _{\rm{1}}}{\rm{ + 1}}}}{{{p}}} \rfloor$). For the convenience of explanation below, we denote the components for extracting the global trend and the periodic trend as
 \emph{ Expert 1} and \emph{ Expert 2}, respectively.
  \begin{figure}
  \centering
  \includegraphics[width=2.6in]{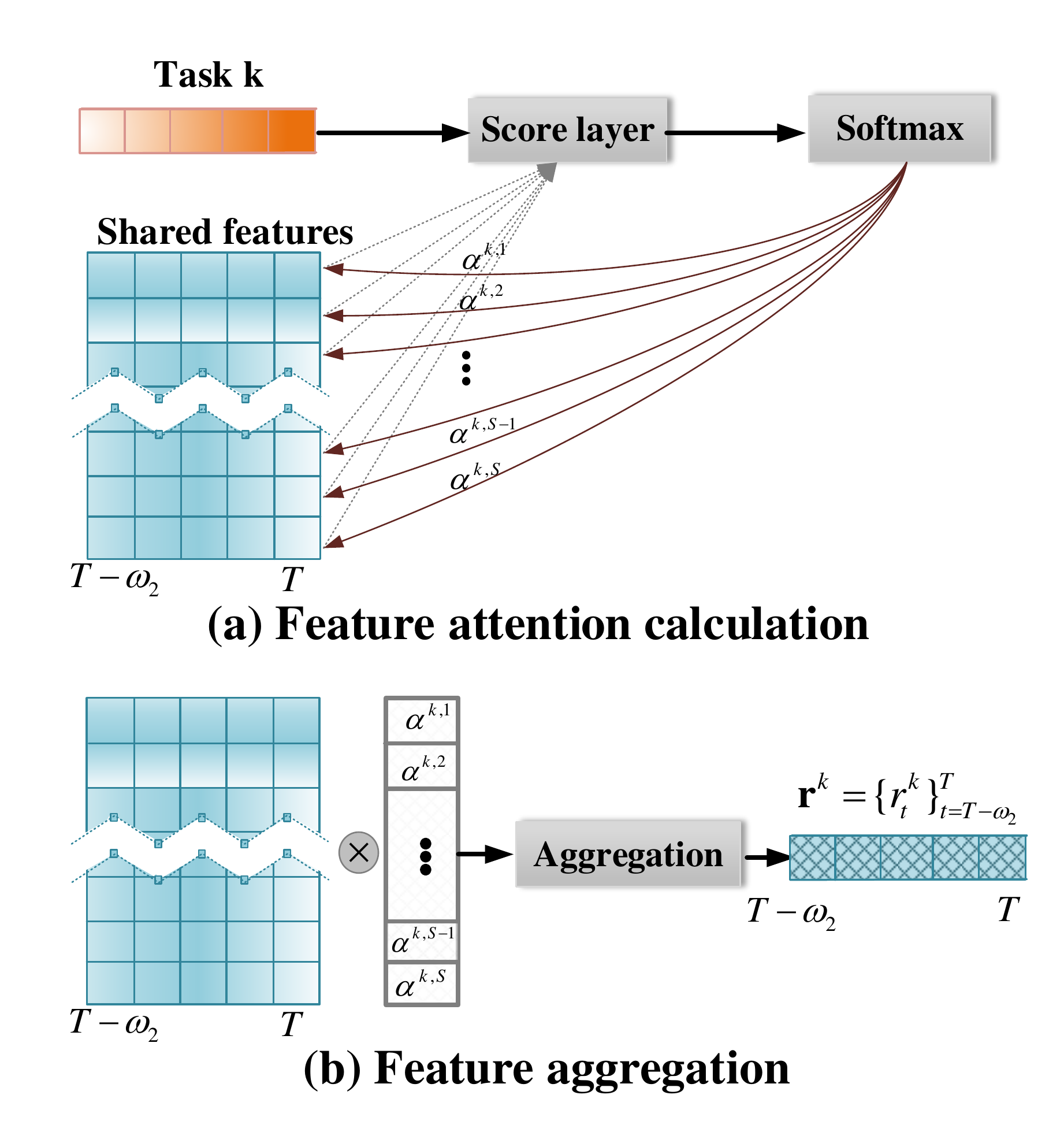}\\
  \caption{The feature selection strategy. (a) Feature attention calculation; (b) Feature selection.}
  \label{figatt}
\end{figure}
 \subsection{Task-specific Module}
This module is responsible for extracting features related to each task and achieving final calibration. The module consists of three parts: a task-specific feature selection strategy, a task tradeoff mechanism based on an extended MMoE, and a calibration unit. We detail each part as follows.
 \subsubsection{Task-Specific Feature Selection}

We designed this feature selection strategy to adaptively retrieve the features most relevant to each task. As illustrated in Fig. \ref{figatt}, the feature selection strategy consists of two parts: feature attention calculation and feature aggregation.

We first introduce the process of feature attention calculation (Fig. \ref{figatt}(a)). This process is to calculate a relevant score for each shared feature regarding to the current task via an attention mechanism. Specifically, given the $k$-th task-specific feature ${\bf{z}}^k$ and the $s$-th shared feature ${\bf{x}}^s$, the attention mechanism calculates the relevant score $\alpha^{k,s}$ of ${\bf{x}}^s$ regarding to ${\bf{z}}^k$ as:
 \begin{equation}
\label{eq7}
 \begin{array}{l}
{e^{k,s}} ={ {\bf{v}}_a^\top}\tanh ({\bf{W}}_a{{\bf{z}}^k} + {\bf{U}}_a{{\bf{x}}^s} + {\bf{b}}_a),\\
{\alpha ^{k,s}} = \frac{{{e^{k,s}}}}{{\sum\nolimits_{s = 1}^{S} { ({e^{k,s}})} }},

\end{array}
\end{equation}
where ${\bf{W}}_a$, ${\bf{U}}_a$, ${\bf{v}}_a$ and ${\bf{b}}_a$ are learnable parameters. 
Then, we aggregate different features with the acquired attention weights, which allow us to dynamically update the impact of each feature on the target task:

\begin{equation}
\label{eq72}
  {{\bf{r}}^k} = \sum\limits_{s = 1}^S {{\alpha ^{k,s}}{{\bf{x}}^s}} ,
\end{equation}
where ${{\bf{r}}^k} = \{ r_t^k\} _{t = T - {\omega _2}}^T$ denotes the aggregated feature that can capture the dynamic changes of the feature relationship at different times for the target task.

After extracting relevant features, we send the selected features ${r}_t^{k}$ and current task-specific feature $z_t^k$ within a time window of length $\omega_2 + 1$ into a GRU to acquire a task-specific trend feature ${\bf{h}}_{T}^k$:  
\begin{equation}\label{eq9}
  {{\bf{h}}_{T}^k} = GRU^{\rm{specific}}([\{ z_t^k\} _{t = T - {\omega _2}}^T;\{ {{\bf{r}}_t^k}\} _{t = T - {\omega _2}}^T]),
\end{equation}
where $[\cdot ; \cdot]$ denotes the concatenation operation.

\begin{figure*}
  \centering
  \includegraphics[width=4.5in]{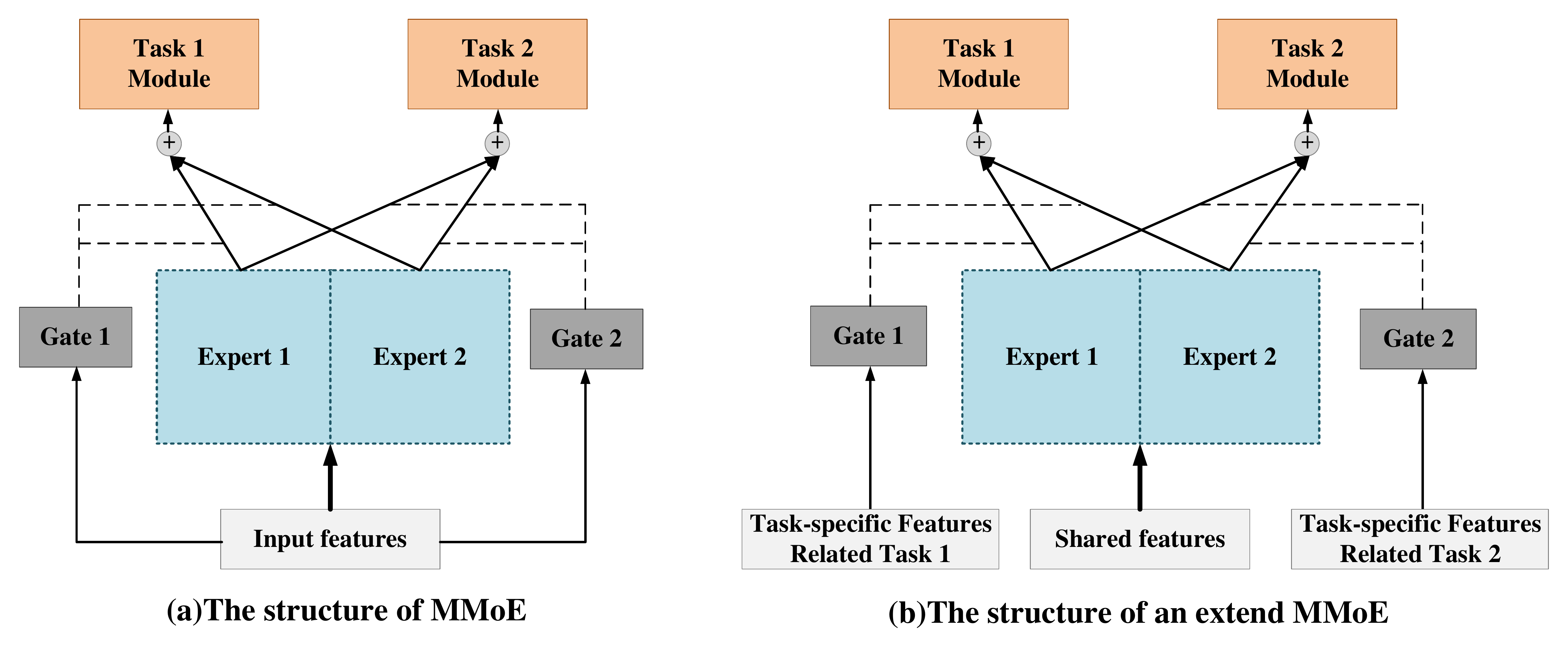}\\
  \caption{The difference between MMoE and extended MMoE.}\label{MMOE}
\end{figure*}
\subsubsection{Task Tradeoffs Based on an Extended MMoE}
As MTNet jointly performs multiple calibration tasks, different tasks pose different demands on the shared trend feature. We model this variability through extending the multi-gate mixture-of-experts (MMoE) \cite{ma2018modeling} structure. MMoE endows the task-specific layer with the ability to flexibly choose the most relevant information from that passed by the shared layers. The original MMoE is characterized by a shared-bottom structure, as shown in Fig.\ref{MMOE} (a).
To satisfy our multi-task scenario, we extend it to a shared-specific architecture, as illustrated in Fig.\ref{MMOE} (b). Specifically, we add gated networks (the left and right sides of Fig.\ref{MMOE} (b)) to enable the communication between each task-specific layer and the shared layer, enhancing the task relevance of the chosen information for each task-specific module. Given a task-specific trend feature ${\bf{h}}_T^{k}$, the gated network first calculates a influence weight vector via a linear transformation with the softmax activation:

\begin{equation}\label{eq9}
  {\bf{g}}^k = softmax ({{\bf{W}}_{gk}}{\bf{h}}_T^k + {{\bf{b}}_{gk}}),
\end{equation}
where $ {\bf{g}}^k $ collects the influence weights of every experts to the $k$-$th$ calibration task. Taking the number of calibration tasks as two as an example, $\mathbf{g}^k=[g_1^k,g_2^k]\in\mathbb{R}^2$. 
Base on the influence weights, the gated network assembles the features of every experts via a weighted summation:
\begin{equation}\label{eq10}
{{\bf{c}}^k} = g_1^k{{\bf{q}}_T} + g_2^k{{\bf{p}}_T}.
\end{equation}
The gated networks can learn mixture patterns of experts, and thus capture the task relationships. After the features are captured by shared module and task-specific module, we concatenate the task-specifc trend ${\bf{h}}_T^k$ and assembly shared feature ${{\bf{c}}^k}$ to form a comprehensive feature:

\begin{equation}
\label{eq11}
{\bf{h}}{{\bf{c}}^k} = [{\bf{h}}_T^k;{{\bf{c}}^k}],
\end{equation}
 where ${\bf{hc}}^k$ encapsulates all trend information regarding to the $k$-th task.

\subsubsection{Calibration Unit}

This unit integrates all relevant information and yields the calibration result. Besides ${\bf{hc}}^k$, we introduce ${\bf{x}}_T$ as the input of this unit, to enhance the impact of the shared features at time $T$ on the result. Specifically, this unit decodes a hidden vector from ${\bf{hc}}^k$ and ${\bf{x}}_T$ through a GRU, and then performs a fully connected operation on the hidden vector to yield the final result $\tilde{y}_T^k$:

\begin{equation}
{{\tilde y}_{T }^k} = FC^{cal}(f({{\mathbf{hc}}^{k}},{{ {\mathbf{x}_T}}}))={\mathbf{w}}_c^ \top f({{\mathbf{hc}}^{k}},{{ {\mathbf{x}_T}}}) + {b^k_c},
\end{equation}
where $f$ represents a GRU cell. ${\mathbf{w}}_c$ and $b_c$ are learnable parameters. 

\section{Experiments}
In this section, we evaluate the performance of the proposed MTNet. All experiments are conducted on a 64-bit Ubuntu 16.04 computer with 18 Intel 2.68GHz CPUs, 256 GB memory, and 8 NVIDIA TITAN X GPUs. All methods have been repeated ten times.

\subsection{Setting}
\begin{table*}
\caption{Shared feature descriptions. }\label{SHARED}
  \renewcommand{\arraystretch}{1.2}
  \centering
  \smallskip\resizebox{0.68\textwidth}{!}{
  \begin{tabular}{c|c|c|c|c}
   \hline
      Dataset&  Features & Data descriptions & \multicolumn{2}{c}{Dimension}\\

     \hline

     \multirow{7}{*} {Fuzhou $/$ Lanzhou dataset} &\multirow{1}{*}{Air pollutants }& CO, O$_3$, NO$_2$, SO$_2$, PM2.5, PM10&6&\multirow{7}{*}{16}\\

       \cline{2-4}
      &\multirow{1}{*}{Weather related features}&Humidity, Temperature&2&\\
      \cline{2-4}
      &\multirow{5}{*}{Empiric features}&PM05N, PM1N, PM25N, PM10N,&\multirow{5}{*}{8}&\\
       & &PM\_ZUFEN= PM25/PM10,&&\\
       & &PMN\_SUM = PM25N+PM10N+PM1N+PM05N,&&\\
      & &PM10N\_ratio = PM10N/PMN\_SUM,&&\\
      & &PM25N\_ratio = PM25N/PMN\_SUM&&\\

  \hline
  \hline

     \multirow{3}{*} {ENEA Pirelli dataset}&  \multirow{1}{*}{Air pollutants }& CO, O$_3$ , NO$_2$, NO$_x$, NMHC&5&\multirow{3}{*}{8}\\

       \cline{2-4}
      &\multirow{1}{*}{Weather related features}&AH$^{\rm a}$,RH$^{\rm b}$,Temperature&3&\\
      \cline{2-4}
      &Empiric features&-&-&\\

      \hline
  \end{tabular}}

  \footnotesize{$^{\rm a}$ refers to absolute humidity.}
\footnotesize{$^{\rm b}$ refers to relative humidity.}\\
\end{table*}

\subsubsection{Datasets}
We conduct our experiments over three datasets, namely Lanzhou dataset, Fuzhou dataset \cite{yu2020airnet,yu2020deep,wang2020category} and ENEA Pirelli dataset \cite{de2018calibrating}.  The data in Lanzhou and Fuzhou datasets are collected by the same type of equipment. These two datasets include the data captured by one micro station and the corresponding static station \footnote{http://www.cnemc.cn/}, which cover measurements from November 2017 to October 2018. For these two datasets, Fuzhou and Lanzhou were chosen as the target regions because their environments are very different as a result of topographic, climatic and social factors. Lanzhou is one of the most polluted cities in China, while the air quality of Fuzhou is relatively good. The ENEA Pirelli dataset came from the source: https://archive.ics.uci.edu/ml/datasets/Air+Quality, which contains the responses of a gas multisensor device. The device was located on the field in a significantly polluted area at road level in an Italian city. Data were recorded from March 2004 to February 2005. Different from the first two datasets, the reference measurement results of ENEA Pirelli dataset were provided by co-located reference certified analyzer. Note that the reference measurement results and sensor data responses used in these three data sets are all hourly average responses. To clearly describe the shared features selected by different data sets, the specific information of the shared features is shown Table \ref{SHARED}. The shared features of the Fuzhou and Lanzhou datasets are from micro stations, and the shared features of ENEA Pirelli dataset are from the air quality chemical multi-sensor device. In order to alleviate the interference of noise data to a certain extent, we perform data cleaning on three datasets, including removing error data, filling missing data, aligning data from micro stations and static stations. Meanwhile, we divide each dataset by a ratio of 8:1:1 for training, validation and testing in chronological order.

\subsubsection{Evaluation Metrics}
We adopt two conventional evaluation metrics to evaluate our model, including symmetric mean absolute percentage error (SMAPE) and mean absolute error (MAE), both of which are widely utilized in regression problem.
%
 We chose them because MAE depends on the numerical scale and SMAPE is not. Therefore, SMAPE is chosen for a more readable evaluation regardless the data scale. Since our method is based on multi-task learning, the above-mentioned single-task indicators cannot measure the total performance of the model in different tasks. Thus, we introduce a global evaluation metric $\overline {{\rm{SMAPE}}}$,  which is calculated by averaging SMAPE values for every tasks.  

\subsubsection{Implementation Details}
The proposed method is implemented with Pytorch. During training, the learning rate and batch size are set to 0.001 and 128, respectively. Model parameters are optimized by Adam \cite{kingma2014adam} optimizer. MTNet jointly trains all tasks by the joint loss function,  which is defined as:
\begin{equation}
L(\Theta ) = \frac{1}{K}\sum\limits_k {\sum\limits_t {L(y_t^k,\tilde y_t^k)} },
\end{equation}
where $\Theta$ denotes all learnable parameters. $L$ is the loss funcation, which is set to Mean Absolute Error (MAE). Meanwhile, we apply the dropout \cite{srivastava2014dropout} with a rate of 0.2 to prevent overfitting. The historical sequence window lengths of the micro station and the static station (i.e.,
$\omega_1$ and $\omega_2$) are set to 167 and 4, respectively. The specific experimental settings are shown in Table \ref{t1}.
\begin{table}[h!]
\caption{The detailed settings of MTNet.}
  \centering
  \resizebox{.95\columnwidth}{!}{
  \begin{tabular}{c|c|c}
   \hline
   Module&Parameter&Value\\
    \hline
    \multirow{4}{*} {Shared module}&Filter size and number of 1st layer CNN&16$\times 3$,25\\
\cline{2-3}
  &Filter size and number of 2nd layer CNN&25$\times 3$,50\\
   \cline{2-3}
   &Stride of CNN &1\\
   \cline{2-3}
   &Hidden dimension of GRU$^{\rm{global}}$&100\\
   \cline{2-3}
   &Hidden dimension of GRU$^{\rm{periodic}}$&100\\
    \cline{2-3}
   \hline
  \multirow{2}{*} {Task-specific module}&Hidden dimension of GRU$^{\rm{specific}}$&210\\
  \cline{2-3}
   &Output dimension of FC$^{\rm{cal}}$&1\\
    \hline
  \end{tabular}}

  \label{t1}
\end{table}

\subsection{Methods for Comparison}
We compare MTNet with the following baselines.

\begin{itemize}
\item \textbf{LR}: Linear regression (LR) is a linear method to learn the mapping from a task-specific reading to a reference result, and it is commonly applied in sensor calibration field. \cite{spinelle2014calibration,spinelle2015field,spinelle2017field}.
    \item \textbf{MLR}: Multivariate linear regression (MLR), which is widely used for sensor calibration  \cite{spinelle2014calibration,spinelle2015field,spinelle2017field}, attempts to learn the mapping from shared features to a reference result.
 \item \textbf{XGB}: Extreme gradient boosting (XGB) proposed by Chen et al. \cite{chen2016xgboost} is widely used in various fields. At the same time, it is one of the commonly used algorithms in many data mining competitions.
 \item \textbf{CCA-LGBM}: It is a category-based calibration approach (CCA) using Light Gradient Boosting Machine (LGBM) \cite{ke2017lightgbm} for calibrating sensor readings \cite{wang2020category}, aiming to learn fine-tuning regression models according to concentration categories rather than a single model. The optimal n\_estimators are selected from [100, 200,...,900, 1000].
\item \textbf{DeepCM}: It is a deep calibration method (DeepCM) \cite{yu2020deep} for low-cost air monitoring sensors. This method formalizes low-cost sensor calibration task as a time series problem, and extracts multi-level sequence feature for assisting calibration. The parameters of DeepCM are set according to the recommendations in the original paper.

  \item \textbf{MMoE}: Multi-gate mixture-of-experts (MMoE) \cite{ma2018modeling} is a multi-task learning architecture, which explicitly learns to model task relations by using gated networks. The parameters of MMoE are set based on the recommendations of the code \cite{keras_mmoe_2018}.

  \end{itemize}
  \begin{figure}
  \centering
  \includegraphics[width=3.5in]{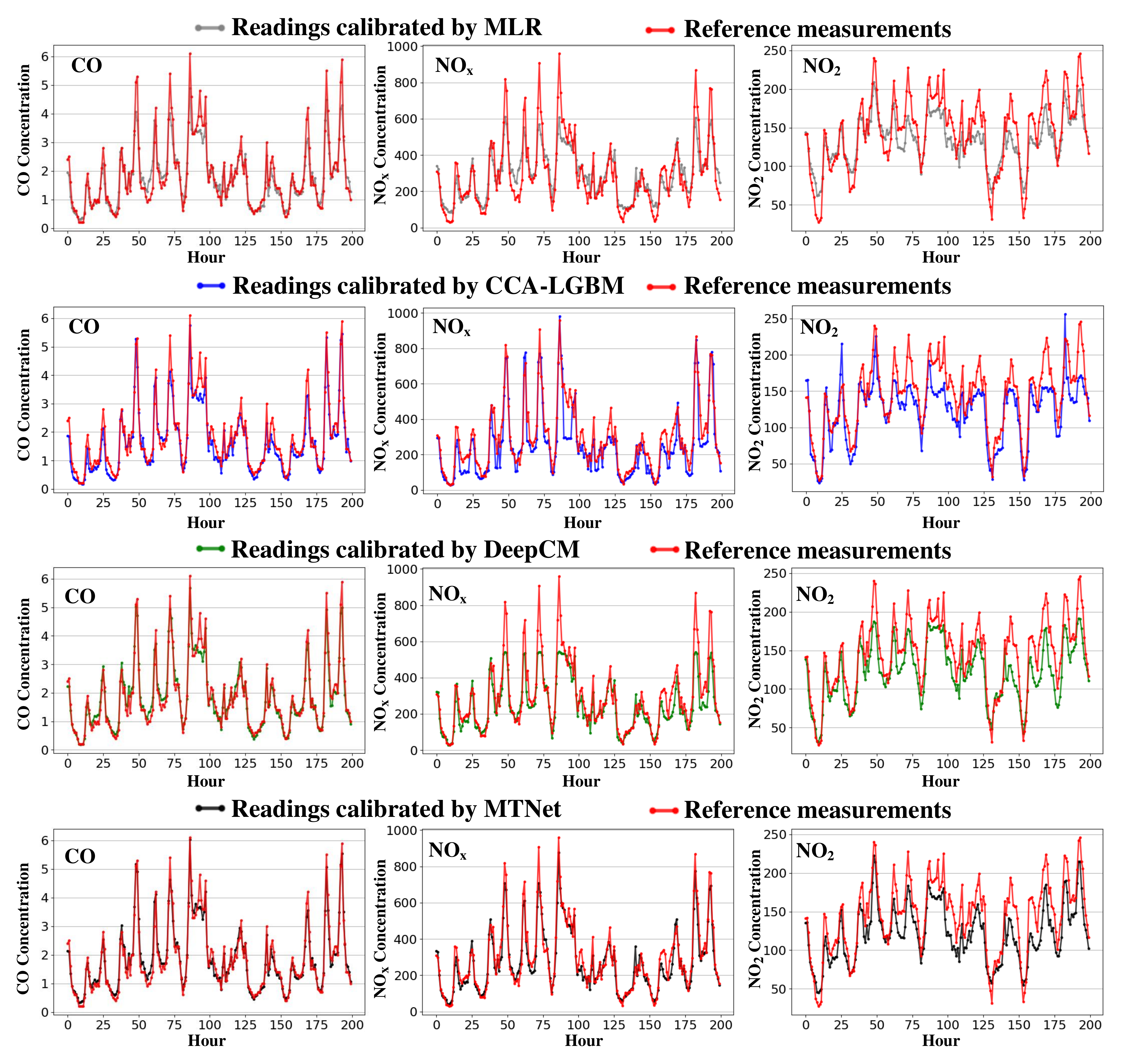}\\
  \caption{Partial calibration results for different methods applied in ENEA Pirelli dataset.}\label{drift}
\end{figure}
   It is worth noting that the input feature set of LR is task-specific feature of sensor to be calibrated, and the input feature sets of MVR, XGB, MMoE and DeepCM are the same as shown in Table \ref{SHARED}. Unlike other methods, the input features of CCA-LGBM are set according to the features recommended by its original paper \cite{wang2020category}.
  Meanwhile, LR, MVR, XGB and CCA-LGBM are implemented by Scikit-Learn. Since these machine learning models do not need to use the validation set to select the best epoch, they are trained by integrating the training set and validation set during training. AirNet \cite{yu2020airnet} is not selected as a baseline because AirNet introduces historical reference measurements as input, which is different from our method in the problem formalization.
\begin{table*}[]
\caption{Experimental results for Lanzhou dataset and Fuzhou dataset. We run each model 10 times and show ``mean $\pm$ standard deviation". }\label{exp1}
\centering
\smallskip\resizebox{0.98\textwidth}{!}{
\begin{tabular}{c|c|c|c|c|c|c|c|c|c|c|c}
\hline
\multirow{3}{*}{Type}& \multirow{3}{*}{Approach} & \multicolumn{5}{c|}{Fuzhou dataset}& \multicolumn{5}{c}{Lanzhou dataset} \\ 
\cline{3-12}
&  & \multicolumn{2}{c|}{CO} & \multicolumn{2}{c|}{NO$_2$} &\multirow{2}{*}{$\overline { {\rm{SMAPE}}} $} &  \multicolumn{2}{c|}{CO} & \multicolumn{2}{c|}{NO$_2$} &\multirow{2}{*}{\textbf{$\overline { {\rm{SMAPE}}} $}}                    \\ \cline{3-6}
\cline{8-11}
                                      &               & SMAPE    & MAE    & SMAPE      & MAE    & &SMAPE    & MAE    & SMAPE      & MAE&                       \\ \hline
\multirow{5}{*}{\textbf{Single-task}} & LR                        & 24.15\%                                 & 0.133                              & 58.92\%                                 & 11.287                             & 41.54\%    & 46.47\%                                 & 0.467                              & 43.39\%                                 & 22.426                              & 44.93\%                    \\ 
                                      & MLR                       & 18.56\%                                 & 0.102                              & 45.02\%                                 & 7.588                              & 31.79\%               & 29.9\%                                  & 0.256                              & 29.52\%                                 & 14.026                              & 29.71\%    \\ 
                                      & XGB                       & 20.03\%                                 & 0.110                              & 36.77\%                                 & 5.854                              & 28.40\%               & 27.14\%                                 & 0.219                              & 25.24\%                                 & 11.520                              & 26.19\%   \\ 
& CCA-LGBM                  & 19.36\%    & 0.105    & 43.28\%     & 7.516      & 31.32\%  &27.74\%&0.223&24.85\%&11.231&26.30\%              \\ 
                                      & DeepCM                    & 19.51\%  $\pm$ 1.06\%  & 0.106 $\pm$ 0.006 & 32.68\% $\pm$ 2.12\%  & 5.281 $\pm$ 0.444 & 26.10\%    & 27.01\% $\pm$ 0.92\%  & 0.211 $\pm$ 0.01  & 25.54\%  $\pm$ 0.77\%  & 11.737 $\pm$ 0.405 & 26.28\%         \\ \hline
\multirow{2}{*}{\textbf{Multi-task}}  & MMoE                      & 19.09\%  $\pm$ 2.84\%  & 0.105 $\pm$ 0.018 & 35.81\%  $\pm$ 2.74\%  & 5.691 $\pm$ 0.630  & 27.45\%    & 28.27\%  $\pm$ 1.56\%  & 0.225 $\pm$ 0.011 &\textbf{ 24.45\%  $\pm$ 0.15\%}  & \textbf{11.154 $\pm$ 0.851} & 26.36\%         \\ 
                                      & MTNet                     & \textbf{16.35\%  $\pm$ 0.9\% }  & \textbf{0.089 $\pm$ 0.005 }& \textbf{32.41\%  $\pm$ 2.00\% } & \textbf{4.946 $\pm$ 0.316} & \textbf{24.38\%}     & \textbf{24.10\%  $\pm$ 0.7\% } & \textbf{0.183 $\pm$ 0.005 }& 26.31\%  $\pm$ 1.10\%  & 12.173 $\pm$ 0.666 &\textbf{ 25.20\%}         \\ \hline
\end{tabular}}
\end{table*}
\begin{table*}[]
\caption{Experimental results for ENEA Pirelli dataset.  We run each model 10 times and show ``mean $\pm$ standard deviation". }
\label{exp2}
\centering
\smallskip\resizebox{0.82\textwidth}{!}{
\begin{tabular}{c|c|c|c|c|c|c|c|c}
\hline
\multirow{3}{*}{Type}  & \multirow{3}{*}{Approach} & \multicolumn{7}{c}{ENEA Pirelli dataset}                                                                                                                                                                                                          \\ \cline{3-9}
&& \multicolumn{2}{c|}{CO}     & \multicolumn{2}{c|}{NO$_x$}     & \multicolumn{2}{l|}{NO$_2$}               &       \multirow{2}{*}{$\overline { {\rm{SMAPE}}} $}                           \\ \cline{3-8}
                                      &                           & SMAPE                                   & MAE                                & SMAPE                                   & MAE                                  & \multicolumn{1}{c|}{SMAPE}              & \multicolumn{1}{c|}{MAE}  &                   \\ \hline
\multirow{4}{*}{\textbf{Single-task}} & LR                        & 37.47\%                                 & 0.631                              & 35.59\%                                 & 90.014                               & 35.30\%                                 & 44.221                              & 36.12\%                \\ 
                                      & MLR                       & 24.19\%                                 & 0.386                              & 21.59\%                                 & 56.155                               & \textbf{ 14.47\%  }                               & \textbf{18.061 }                             & 20.08\%                \\ 
                                      & XGB                       & 27.23\%                                 & 0.417                              & 23.01\%                                  & 58.718                               & 15.01\%                                 & 19.228                              & 21.75\%                \\ 
                                      &CCA-LGBM &22.08\%&0.333&26.65\%&66.650&16.40\%&21.356&21.71\%\\

                                      & DeepCM                    & \textbf{19.00\%  $\pm$  0.51\% } &\textbf{ 0.292 $\pm$ 0.011} & 22.72\%  $\pm$  4.12\%  & 56.821 $\pm$ 10.265 & 19.96\%  $\pm$  2.31\%  & 24.702 $\pm$  2.748 & 20.56\%            \\ \hline
\multirow{2}{*}{\textbf{Multi-task}}  & MMoE                      & 28.88\%  $\pm$  2.35\%  & 0.442 $\pm$  0.036 & 19.60\%  $\pm$  1.30\%  & 47.402 $\pm$  2.906  & 17.67\%  $\pm$ 1.26\%  & 22.339 $\pm$  1.697  & 22.05\%\\ 
                                      & \textbf{MTNet }                    & 20.40\%  $\pm$  1.47\%  & 0.311 $\pm$  0.018 & \textbf{15.70\%  $\pm$  1.62\% } & \textbf{40.768 $\pm$ 4.453  }& 15.32\% $\pm$  0.84\%  & 19.227 $\pm$  1.126& \textbf{17.14\%}            \\ \hline
\end{tabular}}
\end{table*}

\subsection{Baselines}
To demonstrate the effectiveness of our framework, we compare our model with six baselines on three different datasets. Results are shown in Table \ref{exp1} and Table \ref{exp2}. For each dataset, we demonstrate its SMAPE and MAE scores. To give an overall performance evaluation indicator of MTNet, we calculate the average score of SMAPE on multi-task, marked as $\overline {{\rm{SMAPE}}}$. Next, we will analyze the experimental results in detail. Note that for each experiment of the deep learning methods (DeepCM, MMoE and MTNet), due to dropout strategy or the choice of best epoch, the results will be different. Therefore, we run each model 10 times and show the results in the form of "mean $\pm$ standard deviation".

 From Table \ref{exp1} for Fuzhou and Lanzhou dataset, the linear methods (LR and MLR) are generally not effective, especially LR. LR cannot overcome cross-interference, because it only uses its own sensor readings as input features. Although MLR introduces all available features as input, it still performs poorly. This is because the sensor calibration shows a non-linear relationship due to the influence of cross-interference. Non-linear methods (XGB and CCA-LGBM) based on machine learning perform better than linear methods. From the experimental results, it can be seen from the experimental results that, compared with MTNet, the performance of these two models in Fuzhou is significantly lower. This also illustrates the superiority of our multi-task method. The effect of DeepCM, a deep learning method based on a single-task framework, is very competitive, but the overall effect is still slightly inferior to our method.  The performance of MMoE method is also competitive, but the overall performance is inferior to DeepCM and MTNet. One reason is that MMoE does not introduce historical time series. Another reason is that the architecture is not designed based on the characteristics of low-cost sensors.
\begin{table*}[]

\caption{Ablation study results for three datasets. We ran each model 10 times and showed the ablation study results through the  global evaluation metric $\overline {{\rm{SMAPE}}}$. }
\label{exp5}
\centering
\begin{tabular}{c|c|c|c|c|c}
\hline
\multirow{2}{*}{Approach} & Fuzhou dataset & Lanzhou dataset & \multicolumn{3}{c}{ENEA Pirelli dataset} \\ \cline{2-6}
                          & CO-NO$_2$         & CO-NO$_2$          & CO-NO$_x$      & CO-NO$_2$      & CO-NO$_x$-NO$_2$    \\ \hline
MTNet-P                   & 26.06\%        & 25.67\%         & 18.44\%     & 18.73\%     & 18.37\%       \\
MTNet-C                   & 25.76\%        & 26.20\%         & 18.61\%     & 18.80\%     & 17.48\%       \\
MTNet-F                   & 25.55\%        & 26.23\%         & 18.76\%     & 19.04\%     & 18.30\%       \\
MTNet-G                   & 26.12\%        & 25.71\%         & 19.09\%     & 19.11\%     & 17.42\%       \\ \hline
\textbf{MTNet }                    & \textbf{24.38\%}        &\textbf{ 25.20\%  }       & \textbf{18.17\%  }   & \textbf{18.30\%     }& \textbf{17.14\%  }     \\ \hline
\end{tabular}
\end{table*}

The calibration experiment results are also shown in Table \ref{exp2} for ENEA Pirelli dataset. Unlike Fuzhou and Lanzhou, there are two nitrogen oxide sensors (NO$_x$ and NO$_2$) on this dataset. Therefore, for the multi-task learning methods (MMoE and MTNet), a model is needed to simultaneously calibrate the three sensors (CO, NO$_x$ and NO$_2$). It is encouraging that the overall performance of our model has been significantly improved compared to other baseline methods, especially for NO$_x$ sensors. To further demonstrate the rationality of our method, we visualizes the results of three competitive baselines and MTNet in Fig. \ref{drift}. It can be seen from Fig. \ref{drift} that the trends of CO, NO$_x$ and NO$_2$ are very similar, so it is very meaningful to use a multi-task framework to jointly calibrate them to achieve interaction learning of information. In addition, MTNet has greatly improved the calibration of NO$_x$ sensor, especially in the peak position, completing a better fitting. 

%
%
%

Overall, the results
of MTNet are superior to the baseline methods, and MTNet appears suitable for joint calibration of carbon monoxide and nitrogen oxide sensors. 

\subsection{Ablation Studies}
To fully study the performance gained from each component of our proposed model, we implement four other versions of MTNet:
\begin{itemize}
\item \textbf{MTNet-P}: MTNet-P is an incomplete MTNet in which multiple task-specific modules except the calibration units are removed, to verify the influence of the task-specific module. It is worth noting that after the task-specific module is removed, the calibration unit directly receives the assembly shared features of experts ${\bf{c}}^k$ without gated structure guidance, where ${\bf{c}}^k={\bf{q}}_T+{\bf{p}}_T$.
\item \textbf{MTNet-C}: MTNet-C is an incomplete MTNet in which only one convolutional layer remains to test the necessity of introducing the multi-layer convolutional neural network.
     \item \textbf{MTNet-F}: MTNet-F is an incomplete MTNet in which the feature selection strategy is removed in the cause of verifying the significance of feature selection strategy.
\item \textbf{MTNet-G}: MTNet-G is an incomplete MTNet in which the gate operation in the extended MMoE is removed to verify the importance of gate structure. Note that the assembly feature processing method is the same as MTNet-P.

\end{itemize}

Table \ref{exp5} lists the results of four ablation studies to fully investigate the performance of each component of our proposed model. For the ENEA Pirelli dataset, since it contains two nitrogen oxide sensors, we have implemented three groups of multi-task experiments, namely the joint calibration of CO and NO$_x$ sensors, the joint calibration of CO and NO$_2$ sensors, and the joint calibration of CO, NO$_x$ and NO$_2$ sensors. Next, we will discuss the results of ablation studies, and there are several observations worth highlighting: 
(1) The best results on three datasets are  all obtained by MTNet, which demonstrates that every component of our proposed MTNet is effective for joint calibration. (2) MTNet-P removes the task-specific modules in MTNet, and its architecture is actually equivalent to a shared-bottom structure. It can be seen from the experimental results that the overall performance of our model is better than that of MTNet-P, which shows the superiority of our shared-specific architecture. (3) The performance of MTNet-C drops to varying degrees. This shows that multi-task learning is necessary to mine deeper features when mining interaction features using convolution operations. (4) The calibration performance of MTNet-F declines significantly, which indicates that feature selection strategy is essential for task-specific modules. (5) The superiority of the gate structure is proved by comparing MTNet and MTNet-G. In summary, these ablation studies verify the effectiveness of each component of our proposed MTNet.

\section{Conclusions}
This paper presents a new multi-task formulation for low-cost sensor calibration. Beyond traditional single-task calibration frameworks, the new formulation considers the interactions among different tasks and supports to calibrate carbon monoxide and nitrogen oxide sensors simultaneously. With the new formulation, we proposed a multi-task calibration neural network (MTNet).  MTNet exploits a shared module to capture shared interaction information, while using multiple task-specific modules to obtain task-specific information for each task, which proves useful in improving the calibration results. To customize the relevant feature set for each task, we designed a feature selection strategy. Meanwhile, we utilized an extended MMoE strategy to establish communication between the shared module and each task-specific module. Encouragingly, the experimental results on three real-world datasets verify the effectiveness of the proposed new formulation and method.

 There are some topics that we will investigate in the future.  Firstly, we will introduce spatial information besides the temporal features and investigate the sensor calibration problem in a spatiotemporal perspective. Secondly, we will try to deploy MTNet on the web server to provide calibration services for low-cost sensors.


%

%

%
%





\bibliography{mybibfile}
\bibliographystyle{IEEEtran}

\end{document}